\documentclass{article}

\usepackage{PRIMEarxiv}

\usepackage[utf8]{inputenc} 
\usepackage[T1]{fontenc}    
\usepackage{hyperref}       
\usepackage{url}            
\usepackage{booktabs}       
\usepackage{amsfonts}       
\usepackage{nicefrac}       
\usepackage{microtype}      
\usepackage{lipsum}
\usepackage{fancyhdr}       
\usepackage{graphicx}       

\usepackage{multirow}
\usepackage{threeparttable}
\usepackage{bm}
\graphicspath{{media/}}     

\title{graph2topic: an opensource topic modeling framework based on sentence embedding and community detection
}

\author{
  Leihan Zhang\dag, Jiapeng Liu\dag\thanks{Corresponding author. \dag These authors contributed equally to the paper.}, Qiang Yan \\
  School of Modern Post (School of Automation) \\
  Beijing University of Posts and Telecommunications \\
  Beijing, 100876, China\\
  \texttt{zhangleihan@gmail.com} \\
  \texttt{liujiapeng36@foxmail.com} \\
  \texttt{yan@bupt.edu.cn} \\
}

\begin{document}
\maketitle

\begin{abstract}
It has been reported that clustering-based topic models, which cluster high-quality sentence embeddings with an appropriate word selection method, can generate better topics than generative probabilistic topic models. However, these approaches suffer from the inability to select appropriate parameters and incomplete models that overlook the quantitative relation between words with topics and topics with text. To solve these issues, we propose \textit{graph to topic} (G2T), a simple but effective framework for topic modelling. The framework is composed of four modules. First, document representation is acquired using pretrained language models. Second, a semantic graph is constructed according to the similarity between document representations. Third, communities in document semantic graphs are identified, and the relationship between topics and documents is quantified accordingly. Fourth, the word--topic distribution is computed based on a variant of TFIDF. Automatic evaluation suggests that G2T achieved state-of-the-art performance on both English and Chinese documents with different lengths.
\end{abstract}


\section{Introduction}
\label{intro}
Topic modelling is a text-mining method for discovering hidden semantic structures in a collection of documents. It has been widely used outside of computer science, including social and cultural studies\cite{mohr2013}, bioinformatics\cite{liu2010}, and political science\cite{grimmer2013,isoaho2021}. The most popular and classic topic modelling method is latent Dirichlet allocation (LDA)\cite{lda2003}, which provides a mathematically rigorous probabilistic model for topic modelling. The probabilistic model can offer a quantitative expression of the correlation between words with topics and topics with document, which makes it applicable to various quantitative analyses. However, LDA suffers from several conceptual and practical flaws: (1) LDA represents text as bag-of-words, which ignores the contextual and sequential correlation between words; (2) there is no justification for modelling the distributions of topics in text and words in topics with the Dirichlet prior besides mathematical convenience\cite{gerlach2018}; (3) the inability to choose the appropriate number of topics; and (4) the quality of topics, such as coherence and diversity, leaves much to be desired.

Fortunately, contextual embedding techniques provide a new paradigm for representing text and further help alleviate the flaws of conventional topic models, such as LDA. Bidirectional encoder representations from transformers (BERT)\cite{kenton2019} and its variations (e.g., RoBERTa\cite{liu2019}, sentence-BERT\cite{nils2019}, SimCSE\cite{gao2021}), can generate high-quality contextual word and sentence vector representations, which allow the meaning of texts to be encoded in such a way that similar texts are located close to each other in vector space. Researchers have made many fruitful attempts and significant progress in adopting these contextual representations for topic modelling. BERTopic\cite{maarten2022} and CETopic\cite{zhang-cetopic-2022} are the state-of-the-art topic models.

A few studies have suggested that relating the identification of topic structures to the problem of finding communities in complex networks is a promising alternative for topic models\cite{gerlach2018,Lukas2021}. Because the result of community detection algorithms is stable and reliable, which does not rely on any arbitrary parameters, and the overlapping community detection algorithms can describe the relation between topics and documents well, we believe that community detection algorithms are more suitable for identifying topical structures in documents. Therefore, we proposed a simple but effective framework for topic modelling---\textit{graph to topic} (G2T), which incorporates pretrained language models (PLMs) and community detection algorithms. 

The main contributions of our work are as follows:

(1) We proposed a simple but effective framework for topic modelling that can make better use of pretrained language models and stably produce more interpretable and coherent topics.

(2) An open source tool of G2T is publicly available at \url{http://github.com/lunar-moon/Graph2Topic}, and the distillation experiments compared the performances of optional methods in different submodules, which provides a reference to use G2T.

\section{Related Research}
\label{rel}

Although topic models vary in frameworks and approaches, they share the same steps, including text representation and topic modelling.

\subsection{Text Representation}
 The poor representation capability of BoW limited the performance of LDA and its various variants, including ProdLDA\cite{srivastava2017autoencoding} and the neural variational document model\cite{pmlr-v48-miao16}. The boom of pretrained language models initiated by word2vec\cite{mikolov2013efficient} and BERT\cite{devlin2019bert} has brought a comprehensive change in vairous natural language processing tasks inlucding topic modelling.

Many researchers have tried to introduce word embeddings to optimize the performance of generative probabilistic topic models\cite{moody2016mixing,Dieng2020,bianchi2021}. LDA2Vec learns dense word vectors jointly with Dirichlet-distributed latent document-level mixtures of topic vectors and can produce sparse, interpretable document mixtures through a nonnegative simplex constraint\cite{moody2016mixing}. The embedded topic model (ETM) integrates word embeddings with traditional topic models and models each word using a categorical distribution whose natural parameter is the inner product of the word's embedding and the embedding of its assigned topic\cite{Dieng2020}. The combined topic model (CombinedTM) combines BoW with contextual embeddings to feed the neural topic model ProdLDA\cite{bianchi2021}. 

Other researchers have used pretrained language models to encode words, sentences, or documents in semantic space, and the correlation between topics, words, and documents is quantified according to the embedding distance \cite{maarten2022,zhang-cetopic-2022,Suzanna2020tired,Dimo2020top2vec}. Sia \cite{Suzanna2020tired} proposed a topic model that clusters pretrained word embedding and incorporates document information for weighted clustering and reranking topic words. Top2Vec uses joint document and word semantic embeddings to find topic vectors\cite{Dimo2020top2vec}. BERTopic\cite{maarten2022} and CETopic\cite{zhang-cetopic-2022} adopt pretrained language models to encode documents and further find topics using clustering algorithms.

\subsection{Topic Modelling}

The existing topic modelling methods can be broadly divided into two paradigms: Bayesian probabilistic topic models (BPTMs)\cite{wang2021survey} and clustering-based topic models (CBTMs).

\subsubsection{Bayesian Probabilistic Topic Models}

LDA performs the inference using Monte Carlo Markov chain sampling\cite{lda2003} or Gibbs sampling. Neural topic models (NTMs)\cite{pmlr-v48-miao16,srivastava2017autoencoding} incorporate neural components for Bayesian inference, and most NTMs are based on the variational autoencoder (VAE) inference framework\cite{srivastava2017autoencoding}). The VAE adopts an encoder-decoder structure: the encoder trains a neural inference network to directly map a BoW representation of documents into a continuous latent embedding, while the decoder reconstructs the BoW from the latent embedding. ProdLDA\cite{srivastava2017autoencoding} based on the VAE has better performance in topic coherence and computational efficiency than original LDA.

More recently, powerful contextualized embeddings were also incorporated into NTMs, which further improved the topic coherence compared to conventional NTMs that use bag-of-words (BoW) as document representations\cite{bianchi2021,jin-etal-2021-neural}. However, most of the NTMs are based on the variational autoencoder inference framework\cite{kingma2022autoencoding}, which suffers from hyperparameter tuning, computational complexity\cite{zhao2021topic}, and an inability to choose the proper number of topics.

\subsubsection{Clustering-based Topic Models}

Considering the high-quality contextual text representations of pretrained language models and the complexity of the VAE, a few researchers have attempted to produce topics by directly clustering contextual text representations. Clustered-based topic models can generate more coherent and diverse topics than BPTMs\cite{maarten2022,Dimo2020top2vec,zhang-cetopic-2022} with less runtime and lower computational complexity. Text embeddings can be clustered to model the correlation between words with topics and topics with documents. Both clustering and community detection can fulfil the task. The SOTA topic models (BERTopic\cite{maarten2022} and CETopic\cite{zhang-cetopic-2022}) use centroid-based and density-based clustering algorithms such as K-means and HDBSCAN to cluster documents. However, these topic models have the following defects: the parameters of clustering algorithms usually need to be manually assigned, which results in the instability of topic models; the appropriate topic number is hard to choose; and both BERTopic and CETopic implicitly assume that one document contains only one topic, which conflicts with the fact that long documents usually contain more than one topic.

\section{Proposed Method}

\subsection{Notation and Terminology}

We follow notations and definitions similar to those of the original LDA\cite{lda2003}. The topic model is used to extract topics from a \textit{corpus} or a \textit{document}, a corpus consists of a set of \textit{documents}, and a document is made up of a sequence of \textit{words}. The basic assumption is that documents are represented as mixtures over latent topics, where each topic is characterized by a distribution over words\cite{lda2003}. A document can contain one or many topics.
\begin{itemize}
  \item A \textit{word} is the basic unit of text and can be defined as an item from a vocabulary indexed by $\{1,2,...,V\}$.
  \item A \textit{document} is a sequence of $N$ words and can be represented by $d = (w_1,w_2,...,w_N)$, where $w_n$ is the $n_{th}$ word in the sequence.
  \item A \textit{corpus} is a set of $M$ documents denoted $D=\{d_1,d_2,..,d_M\}$, where $d_m$ is the $m_{th}$ document in the corpus.
  \item A \textit{topic} is composed of a set of words with weight or probability. Let $\beta$ be the distribution of words in a topic, which can be denoted $\beta=\{(w_j,weight_j)|j\in[1,J]\}$, where $\sum_{j=1}^Jweight_j=1$. 
  \item For the \textit{topics} $T=\{t_k|k\in[1,K]\}$ extracted from a document or a corpus, let $\alpha$ be the distribution of topics in a document or a corpus, which can be denoted $\alpha=\{(t_k,weight_k)|k\in[1,K]\}$, where $\sum_{k=1}^Kweight_k=1$.
\end{itemize}

\subsection{Graph to Topic}
\begin{figure}
\centering
\includegraphics[width=\textwidth]{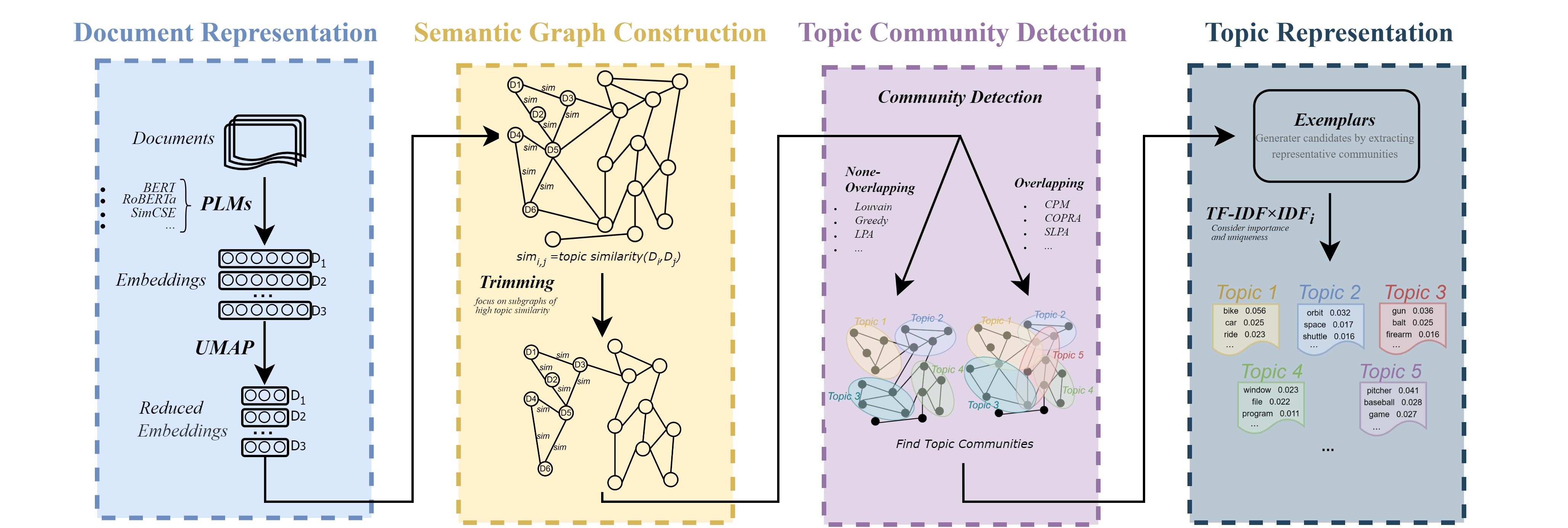}
\caption{The framework of G2T}
\label{fig2}
\end{figure}

Based on pretrained language models and community detection algorithms, we propose a simple but effective topic modelling framework---document semantic similarity \textit{\underline{g}raph to \underline{t}opics} (G2T for short). As shown in Figure \ref{fig2}, G2T is composed of four submodules: document representation, semantic graph construction, topic community detection, and topic representation.  

\subsubsection{Document Representation}

Since topic modelling mainly deals with sentence-level semantics and sentence-level embedding can capture much more integrated contextual information than vocabulary-level embedding, we adopt sentence embeddings to represent documents.

A series of methods can encode sentences into embeddings. Doc2vec can represent variable-length pieces of texts, such as sentences, paragraphs, and documents, with fixed-length feature representations\cite{Le2014doc2vec}. However, its effectiveness is limited by placing the same weights on all the words in a document. To solve this issue, researchers have attempted to obtain sentence embeddings by averaging the BERT embeddings or using the output of the [CLS] token. Because these methods tend to induce a nonsmooth anisotropic semantic space of sentences, their performance on semantic similarity computation is even worse than that of averaging GloVe embeddings\cite{pennington2014glove,nils2019}. For ease of use, Reimers\cite{nils2019} proposed sentence-BERT, a modification of the pretrained BERT network, which uses Siamese and triplet networks to derive semantically meaningful sentence embeddings. The embedding performs well in semantic comparison using cosine similarity. Postprocessing methods such as BERT-flow\cite{li-etal-2020-sentence} and BERT-whitening\cite{su2021whitening} can also significantly improve the uniformity of the semantic space of sentences, but they degenerate the alignment of expected paired sentences. To solve this problem, Gao\cite{gao2021} presented a simple contrastive learning framework---SimCSE for sentence embeddings, which has state-of-the-art performance on tasks using sentence embeddings. The contrastive learning objective can regularize the pretrained embedding' anisotropic space to be more uniform and better align similar sentence pairs.

In addition to the representation quality of different embedding methods, reducing the dimension of sentence embeddings can improve the computational efficiency and maintain the effectiveness\cite{Dimo2020top2vec,maarten2022,zhang-cetopic-2022}. The semantic information in an embedding is redundant to some extent for topic modelling. Above all, reducing the dimension of an embedding might be a good option.UMAP is a nonlinear dimension reduction method and can strike a balance between emphasizing the local versus global structure of initial data\cite{mcinnes2018umap}.

\subsubsection{Semantic Graph Construction}
Gerlach\cite{gerlach2018} represented a word--document matrix as a bipartite network and performed a nonparametric Bayesian parametrization on pLSI from a community detection method---hierarchical SBM, which can match the statistical properties of real text better than LDA. GraphTMT\cite{Lukas2021} uses a community detection method to cluster word embeddings and can produce much more coherent and meaningful topics than LDA and these clustering-based topic models. Thus, we believe that community detection on semantic graphs of document-level embeddings can capture many more integrated semantics than word embeddings and can further help build better topic models.

Since sentence embedding methods such as SimCSE and sentence-BERT can encode text as long as 512 tokens, we use sentence embeddings to represent documents. For a document $d_i$ in corpus $D$, its embedding can be represented by $\bm{e_i}$. The similarity $sim(d_i,d_j)$ between any two sentences $d_i$ and $d_j$ can be calculated by the cosine similarity of $\bm{e_i}$ and $\bm{e_j}$, as shown in Eq (\ref{eq0}). Then, the semantic graph can be represented by an undirected weighted graph $G=(\mathbb{N}, \mathbb{E})$, where each node represents a document and the similarity between each pair of documents represents the weight of the edge between the documents. Except for the cosine similarity, a few other similarity computation methods can be alternatives, and their effects on G2T are investigated in the distillation experiments.

\begin{equation}
\label{eq0}
sim(d_i,d_j)=\frac{\bm{e_i} \cdot \bm{e_j}}{|\bm{e_i}| \cdot |\bm{e_j}|}
\end{equation}

To reduce the complexity of community detection, the undirected weighted complete graph is pruned and transformed into an undirected unweighted incomplete graph. Specifically, the edges with relatively small weights are removed. Considering Zipf's law\cite{Andrade1936} in natural language and that most similarities between document embeddings are higher than 0.9, we reserve the edges with the top-$P$ percent of weights. After removing a portion of edges, isolated nodes arise. These isolated nodes are identified as trivial topics among the corpus and are ignored. Then, community detection is performed on the maximum connected subgraph $G_s$.  

\subsubsection{Topic Community Detection}

A document may contain one or more topics. When a document contains only one topic, each node in $G_s$ can only belong to one community. When a document contains more than one topic, the node in $G_s$ can belong to more than one community. The two cases can be solved by non-overlapping and overlapping community detection algorithms. The non-overlapping algorithms include LPA\cite{pre2017}, greedy modularity\cite{pre2004}, Louvain\cite{Blondel_2008}, and k-components\cite{James-2003-kcomponents}. The overlapping algorithms include COPRA\cite{Gregory_2010}, SLPA\cite{xie2011slpa}, the CPM\cite{palla2005uncovering}, and the LFM\cite{Lancichinetti_2009}. These algorithms can automatically determine the number of communities, and some of them can even find hierarchical community structures, which is important for topic modelling. The effects of these algorithms on G2T are also compared.

After community detection, topic communities $C=\{c_k|k\in[1,K]\}$ are identified, and each community corresponds to a document set $D_k$, where $D_k=\{d_{km}|m\in[1,M^{'}]\} \subseteq D$. Each community $c_k$ is used to produce a topic $t_k$. The topic representation is introduced in the next section. The distribution of topics $\alpha=\{(t_k,weight_k)|k\in[1,K]\}$ in document $d_i$ is derived by the softmax probability of the cosine similarity of $d_i$ with $D_k$ among all topics (Eq (\ref{eq1})), and the cosine similarity $CS(d_i, D_k)$ of $d_i$ with $D_k$ is approximately represented by the average similarity of $d_i$ with the most similar 10 documents $D_k^i$ in $D_k$. For non-overlapping community detection, the distribution $\alpha$ degrades to a one-hot representation.
\begin{equation}
CS(d_i, D_k) = \frac{1}{10}\sum_{i\neq j}^{d_j\in D_k^i} cosine(v_i,v_j)
\end{equation}
\begin{equation}
\label{eq1}
\alpha(weight_k) = \frac{exp(CS(d_i, D_k)} {\sum_{k=1}^Kexp(CS(d_i, D_k))}
\end{equation}

\subsubsection{Topic Representation}
Selecting representative words from the clustered documents matters greatly for the interpretability and coherence of topics. We adopt the topic word selection method proposed by Zhang\cite{zhang-cetopic-2022}, which incorporates global word importance with term frequency across clusters. Global word importance is measured by the classic TFIDF. Let $c_{w_n,d_m}$ be the frequency of word $w_n$ in document $d_m$, $\sum_{m=1,n=1}^{M,N} {c_{w_n,d_m}}$ be the total word' frequency in the corpus $D$, and $df_{w_n}$ be the number of clusters in which word $w_n$ appears. In Eq (\ref{eq2}), TFIDF marks the words that important to each document in the entire corpus, while $IDF_i(w_n)$ penalizes the common words in multiple clusters. Let $N'$ be the number of words representing a topic, and the weight or probability $\beta(weight_k(w_n))$ of $w_n$ in topic $t_k$ can be computed by the softmax function of $TFIDF \cdot IDF_i(w_n)$.

\begin{equation}
\label{eq2}
TFIDF \cdot IDF_i(w_n) =\frac{c_{w_n,d_m}}{\sum_{m=1,n=1}^{M,N} c_{w_n,d_m}} \cdot \frac{K}{df_{d_m}} 
\end{equation}

\section{Experiments and Results}
\label{exp}
\subsection{Datasets}

Five English datasets and two Chinese datasets were used to evaluate the performance of G2T. The English datasets include 20 NewsGroups, BBC News, M10, CitySearch, and Beer, and the Chinese datasets are ASAP and the dataset of the NLPCC-2017 shared task. As the number of Chinese datasets is limited and we want to validate the performance of G2T on both short and long documents, the abstracts and contents of NLPCC were used separately as two datasets, i.e., NLPCC-A and NLPCC-C. The statistics of the eight datasets are shown in Table \ref{tab1}.

\textit{20 Newsgroups (20NG)} is a collection of approximately 20,000 newsgroup documents partitioned across nearly 20 different newsgroups\cite{Lang95}.

\textit{M10} is based on 10,310 publications from 10 distinct research areas, including agriculture, archaeology, biology, computer science, financial economics, industrial engineering, material science, petroleum chemistry, physics and social science\cite{lim2015}. The query words of each publication for these ten disciplines are chosen as the dataset.

\textit{BBC News (BBC)} contains 2225 documents from the BBC News website between 2004 and 2005\cite{Greene2006}.

\textit{CitySearch (CRR)} contains 52264 reviews of 5531 restaurants. These reviews consist of an average of 5.28 sentences\cite{Ganu2009}.

\textit{Beer} was collected from the rating websites BeerAdvocate and RateBeer, and the ratings are given in terms of feel, look, smell, taste, and so on\cite{McAuley2012}.

\textit{ASAP} is a review-level corpus and concentrates on the restaurant domain. The 18 aspect categories of ASAP include service\#hospitality, price\#level, ambience\#decoration, and so on\cite{bu2021asap}.

\textit{NLPCC} contains Chinese news articles, and the short summary is used for news browsing and propagation on Toutiao.com\footnote{http://tcci.ccf.org.cn/conference/2017/taskdata.php}.

\newcommand{\tabincell}[2]{\begin{tabular}{@{}#1@{}}#2\end{tabular}}  
\begin{table*}
\small
\centering
\begin{threeparttable}
\caption{Statistics of datasets.}
  \label{tab1}

   \begin{tabular}{l|cccccc}
  \toprule
    Dataset & $\#documents$ & $length\_ave$ &$length\_std$ & $\#vocabulary$ & $\#label$ & $language$ \\
    \midrule
    20Newsgroup  &18 309 &48.01 &130.53 &1 612 &20 &English \\
    M10  &8 355 &5.91 &2.31 &1 696 &10 &English \\
    BBC News &2 225 &120.12 &72.45 &2 949 &5 &English\\
    CitySearch &31 490 &7.18 &4.79 &14 448 &6 &English\\
    Beer &39 216 &6.37 &3.96 &10 378 &5 &English\\
    ASAP &46 730 &212.61 &87.77 &174 955 &--&Chinese\\
    NLPCC-A &20 333 &35.09 &7.92 &77 491 &--&Chinese\\
    NLPCC-C &20 333 &112.34 &35.33 &25 1636 &--&Chinese\\
  \bottomrule
\end{tabular}
\begin{tablenotes}
        \item[1] For each corpus, $\#documents$ represents the number of documents, $length\_ave$ and $length\_std$ denote the mean and standard deviation of the lengths of documents, respectively, $\#vocabulary$ represents the vocabulary size, and $\#label$ represents the number of human-annotated topic labels.  
      \end{tablenotes}
\end{threeparttable}
%
\end{table*}

\begin{figure}
\centering
\includegraphics[width=0.8\textwidth]{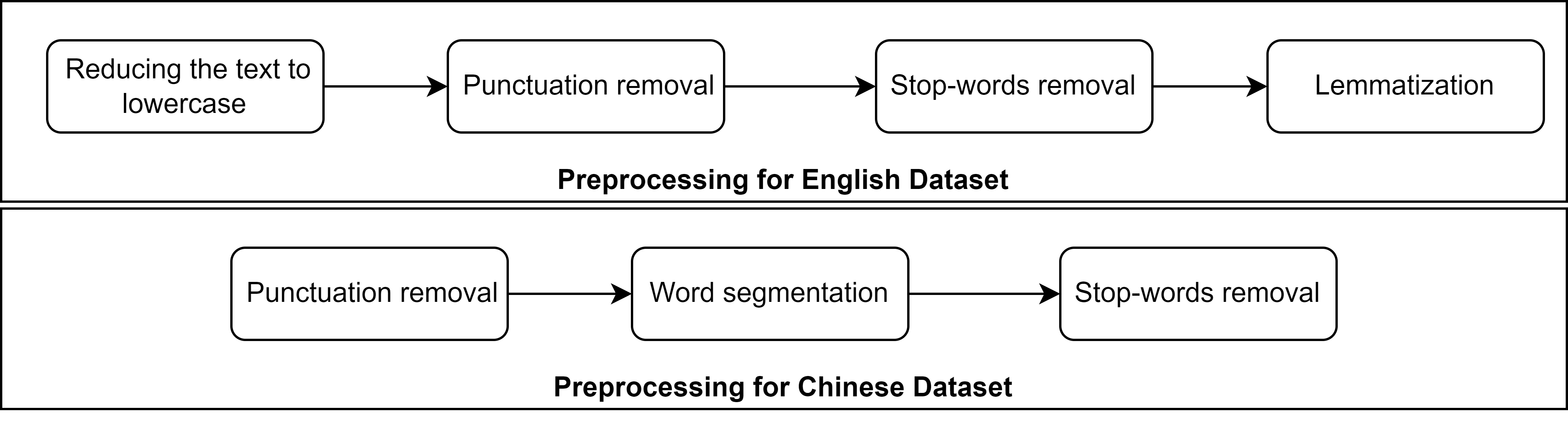}
\caption{Preprocessing flow of rare documents.}
\label{fig3}
\end{figure}

We followed the framework of OCTIS\cite{terragni-etal-2021-octis} to preprocess the original documents. As shown in Figure \ref{fig3}, the English datasets were preprocessed by dropping punctuations, lemmatization, filtering stopwords, and removing documents with less than 5 words; the Chinese datasets were preprocessed by word segmentation and removing punctuation and stopwords. The English stop-words list was acquired from MALLET\footnote{https://mallet.cs.umass.edu/download.php}. The POS lemmatizer of spaCy\footnote{https://spacy.io/} was used to obtain the lemma of words. Chinese documents were segmented using Jieba\footnote{https://github.com/fxsjy/jieba}.

\subsection{Baseline and Parameters}

\label{bas}

Five baseline approaches were chosen to compare with G2T, and these baselines include LDA\cite{lda2003}, the ETM\cite{Dieng2020}, CombinedTM\cite{bianchi2021}, BERTopic\cite{maarten2022}, and CeTopic\cite{zhang-cetopic-2022}.

LDA\cite{lda2003} is the most popular and classic topic model that generates topics via document-topic and topic-word distributions. We used the default parameters for LDA provided by Gensim\footnote{https://radimrehurek.com/gensim/models/ldamodel.html} to extract topics. 

The ETM\cite{Dieng2020} is a generative model of documents that marries traditional topic models with word embeddings.
The default parameters and settings in the original paper were used for the experiment. 

CombinedTM\cite{bianchi2021} extends ProdLDA by concatenating the contextualized sentence-BERT (sBERT for short) \cite{nils2019} embeddings with the original BoW as the new input to feed into the VAE framework. The default parameters in the original paper are used for the experiment. 
BERT-base provided by HuggingFace was used to obtain contextual embeddings.

BERTopic\cite{maarten2022} is a clustering-based method that first leverages the HDBSCAN\cite{mcinnes2017hdbscan} to cluster BERT embeddings of sentences. BERT-base provided by HuggingFace was used to obtain contextual embeddings. And the embedding dimension was reduce to 5 by using UMAP\cite{mcinnes2018umap}.

CETopic\cite{zhang-cetopic-2022} is another clustering-based method that uses UMAP to reduce the embedding dimension and k-means to cluster the reduced-dimension contextual embedding. Cetopic uses unsupervised SimCSE (BERT-base) for English datasets and uses unsupervised SimCSE (RoBERTa-base) for Chinese datasets. In the experiment, the same parameters and methods in the original paper were used. 

Consistent with the baseline, G2T uses unsupervised SimCSE (BERT-base) for English datasets and unsupervised SimCSE (RoBERTa-base) for Chinese datasets. The embedding dimension was reduced to 5 using UMAP. $P$ was set to 95\%. Greedy modularity was used to detect the topic community. 

\subsection{Evaluation Methods}

(1) \textbf{Topic Diversity}

Topic diversity is the percentage of unique words for all topics\cite{Dieng2020}. It ranges from 0 to 1, and a higher percentage means more repeating topics. Let $TD(k)$ represent the diversity of topic $k$, which can be computed by Eq (\ref{eq3}). 

\begin{equation}
\label{eq3}
TD_k = \frac{1}{N}\sum_{n=1}^{N}\frac{1}{f_k(n)}
\end{equation}

\begin{equation}
\label{eq4}
TD = \frac{1}{K}\sum_{k=1}^{K}TD_k
\end{equation}

where $N$ is the number of words constituting a topic and $f_k(n)$ is the occurrence frequency of the words of topic $k$ in all topics. Then, the averaged $TD$ is used to quantify the diversity of the $K$ topics for each topic model. A higher $TD$ value means more diverse topics.

(2) \textbf{Topic Coherence}

It has been proven that the coherence metrics based on both pointwise mutual information (PMI) and the distance between distributed representations of topic words can well emulate human judgement with reasonable performance\cite{lau2014machine}.
Thus, topic coherence was evaluated by using pointwise mutual information-based metrics, including $NPMI$ \cite{bouma2009normalized} and $C_v$\cite{Roder2015}, and the distance-based metric $Q_{w2v}$\cite{Nikolenko2016}.

$NPMI$ refers to the normalized pointwise mutual information of all word pairs of the given topic words that co-occur in the same sliding widow\cite{bouma2009normalized}. For a given ordered set of top words $W=\{w_i| i\in[1,N]\}$ of topic $k$, the NPMI of topic $k$ can be calculated by Eq (\ref{eq5}), and $\epsilon$ is used to avoid a logarithm of zero. Then, the averaged NPMI of the $K$ topics is used to quantify the coherence Eq (\ref{eq6}), and a higher $NPMI$ value means more coherent topics.

\begin{equation}
\label{eq5}
NPMI_k=\sum_{i<j}\frac{log\frac{(P(w_i,w_j)+\epsilon)}{P(w_i) \cdot P(w_j)}}{-log(P(w_i,w_j)+\epsilon)}
\end{equation}

\begin{equation}
\label{eq6}
NPMI=\frac{1}{K}\sum_{k=1}^{K}NPMI_k
\end{equation}

$C_v$ is based on a sliding window, one-set segmentation of the top words and an indirect confirmation measure that uses normalized pointwise mutual information (NPMI) and cosine similarity\cite{Roder2015}. $v_i$ represents the vector of $w_i$, which is constituted by the NPMI of $w_i$ with all words in the same topic $k$. Then, the $C_{vk}$ of topic $k$ can be computed by Eq (\ref{eq8}). The $C_v$ of the $K$ topics computed by Eq (\ref{eq9}) can be used to quantify the coherence, and a higher $C_v$ value means more coherent topics.
\begin{equation}
\label{eq7}
\vec{v_i}=(v_{i1},v_{i2},...v_{in},...,v_{iN})
\end{equation}

\begin{equation}
\label{eq8}
C_{vk}=\sum_{i<j}cos(\vec{v_i},\vec{v_j})
\end{equation}

\begin{equation}
\label{eq9}
C_v=\frac{1}{K}\sum_{k=1}^{K}C_{vk}
\end{equation}

Nikolenko\cite{Nikolenko2016} proposed $Q_{w2v}$ based on the assumption that top words in a topic should be close to each other in semantic space. For a set of top words $W=\{w_i, i\in[1,N]\}$ for topic $k$ with weights, the distributed representation of $w_i$ is $v_i \in \mathbb{R}^d$. As shown in Eq (\ref{eq11}), the quality $Q_{w2v}^k$ of a topic $t_k$ is quantified by the average distance between its topic words. The quality $Q_{w2v}$ of the $K$ topics, which can be computed by Eq (\ref{eq12}), represents the quality of the topics, and smaller values suggest better topics. 

\begin{equation}
\label{eq10}
distance(v_i,v_j) = \sum_{l=1}^{d}(v_{il}-v_{jl})^2
\end{equation}

\begin{equation}
\label{eq11}
Q_{w2v}^k = \frac{1}{N(N-1)}\sum_{i \neq j, i,j \in [1,N]} distance(v_i,v_j)
\end{equation}

\begin{equation}
\label{eq12}
Q_{w2v}=\frac{1}{K}\sum_{k=1}^{K}Q_{w2v}^k
\end{equation}                   

\section{Results}

\begin{table}[]
\caption{Comparison with baselines}
\label{tab2}
\centering
\small
\begin{tabular}{llcccccc}

\toprule
Datasets & Methods & \multicolumn{1}{c}{LDA} & ETM & CombinedTM & \multicolumn{1}{c}{BERTopic} & CETopic & G2T \\ \midrule
\multirow{4}{*}{20NG} & TD   & 0.70 & 0.36 & 0.80 & \textbf{1} & 0.74 & 0.78 \\
                      & NPMI & 0.05 & 0.04 & 0.05  & 0.02 & 0.14   & \textbf{0.18} \\
                      & $C_v$ & 0.53  & 0.50 & 0.58 & 0.40 & 0.70 & \textbf{0.79} \\
                      & $Q_{w2v}$ & 33.43  & 32.09 & \textbf{22.91} & 39.77 & 27.38 & 22.69\\ \hline
\multirow{4}{*}{BBC}  & TD   & 0.31 & 0.25 & 0.61 & 0.76 & 0.80 & \textbf{0.81} \\
                      & NPMI & -0.01 & 0.02 & 0.02 & 0.10 & 0.16  & \textbf{0.17} \\
                      & $C_v$ & 0.36 & 0.42 & 0.64 & 0.67 & 0.81 & \textbf{0.82} \\ 
                      & $Q_{w2v}$ & 21.93 & 13.79 & 6.61 & 8.37 & 5.72 & \textbf{5.52}\\ \hline
\multirow{4}{*}{M10}  & TD   & 0.64 & 0.16 & 0.67 & 0.80 & \textbf{0.87} & 0.84 \\
                      & NPMI & -0.15 & -0.01 & -0.08 & -0.02 & 0.21  & \textbf{0.24} \\
                      & $C_v$ & 0.33 & 0.32 & 0.43 & 0.57 & 0.76 & \textbf{0.81} \\ 
                      & $Q_{w2v}$ &  0.39 &  \textbf{0.09}  & 0.38 &  0.78 &  0.65 & 0.87\\ \hline
\multirow{4}{*}{CRR}  & TD   & 0.85 & 0.17 & 0.72 & 0.68 & 0.70 & \textbf{0.79} \\
                      & NPMI & -0.04 & -0.01 & -0.02 & -0.01 &   & \textbf{0.13} \\
                      & $C_v$ & 0.38 & 0.40 & 0.53 & 0.44 & 0.55 & \textbf{0.72} \\ 
                      & $Q_{w2v}$ & 7.78 & 10.15 & \textbf{5.23} & 6.69  &  7.30 & 6.46\\ \hline
\multirow{4}{*}{Beer} & TD   & \textbf{0.80} & 0.20 & 0.63 & 0.58 & 0.66 & 0.68 \\
                      & NPMI & -0.01 & -0.03 & 0.04 & 0.02 & 0.143 & \textbf{0.144} \\
                      & $C_v$ & 0.47 & 0.36 & 0.61 & 0.44 & 0.65 & \textbf{0.74} \\ 
                      & $Q_{w2v}$ & 8.72 & 18.31 & \textbf{5.57} & 12.21 & 7.90 & 6.65\\ \hline
\multirow{4}{*}{ASAP} & TD   & 0.56 & 0.05 & \textbf{0.88} & 0.77 & 0.48 & 0.83 \\
                      & NPMI & -0.08 & 0.01 & -0.14 & -0.01 & 0.06  & \textbf{0.07} \\
                      & $C_v$ & 0.40 & 0.39 & 0.42 & 0.51 & 0.50 & \textbf{0.71} \\ 
                      & $Q_{w2v}$ & 81.03 & 117.09 & 67.11 & 97.91 & 103.84  & \textbf{52.55}\\ \hline
\multirow{4}{*}{NLPCC-A}  & TD   &0.81 & 0.15 & 0.92 & 0.87 & 0.93 & \textbf{0.94} \\
                      & NPMI & -0.11 & -0.07 & 0.04 & 0.04 & 0.16  & \textbf{0.26} \\
                      & $C_v$ & 0.44 & 0.36 & 0.54 & 0.76 & 0.89 & \textbf{0.95} \\ 
                      & $Q_{w2v}$ & 16.00 & 43.09 & 9.23 & \textbf{9.47} &  12.50 & 12.19\\ \hline
\multirow{4}{*}{NLPCC-C}  & TD   & 0.63 & 0.31 & 0.93 & 0.49 & 0.67 & \textbf{0.95} \\
                      & NPMI & 0.03 & 0.01 & 0.02 & 0.04 & 0.12 & \textbf{0.18} \\
                      & $C_v$ &  0.50 & 0.43 & 0.63 & 0.52 & 0.71 & \textbf{0.89} \\ 
                      & $Q_{w2v}$ & 284.36 & 354.40  &  115.43& 287.57 & 221.44  & \textbf{82.40}\\ \bottomrule
\end{tabular}%
\end{table}

For the methods that require predetermining the number of topics $K$, we computed all the metrics of five groups of topics. The topic number of the five groups of topics was chosen from $\{10,20,30,40,50\}$. For the approaches that do not predetermine the topic number, we computed all the metrics for the top-$Min\{K,50\}$ topics. Then, the performance of each approach was measured using the average metric scores of all groups of topics. Each topic was represented by the top-10 words.

Let $R_k$ be the rate of G2T obtaining the $top\mbox{-}k$ scores on the particular set of metrics among all approaches on a set of datasets. Therefore, $R_1$ is the rate of obtaining the highest scores on a set of metrics across the given datasets, and $R_2$ is the rate of obtaining the $top\mbox{-}2$ scores. 

Table \ref{tab2} shows that G2T generally obtained consistent outstanding coherence and diversity scores on all datasets. Among all approaches on all datasets, $R_1$ and $R_2$ are 23/32 (71.88\%) and 30/32 (93.75\%), respectively. For topic coherence, $R_1$ and $R_2$ are 19/24 (79.16\%) and 22/24 (91.67\%), respectively. For topic diversity, $R_1$ and $R_2$ are 4/8 (50.00\%) and 8/8 (100.00\%), respectively. For different languages, the $R_1$ and $R_2$ of G2T are 13/20 (65\%) and 19/20 (95\%) on English datasets and 13/15 (86.67\%) and 14/15 (93.33\%) on Chinese datasets, respectively. For documents with different lengths, $R_1$ and $R_2$ are 11/12 (91.67\%) and 12/12 (100\%) for long documents and 12/20 (60\%) and 18/20 (90\%) for short documents, respectively. From the above comparison, we can conclude that G2T performs consistently better than baselines on both the English/Chinese corpus and long/short documents. 

For different paradigms of topic models, the pretrained word/document embedding-based approaches, including G2T, CETopic, BERTopic and CombinedTM, extracted higher-quality topics than the traditional probability generation-based approaches such as LDA and the ETM; the topic models, including G2T, CETopic, and BERTopic, which extract topics using clustering methods or community detection algorithms based on high-quality contextualized word/document representations, can obtain more coherent and diverse topics than the ETM and CombinedTM, which incorporate word embeddings into neural topic models (NTMs). Above all, compared with those of SOTA topic models such as BERTopic and CETopic, the diversity and coherence of topics extracted by G2T are much better, which suggests that community detection can better use contextualized embeddings.

\section{Discussion and Conclusion}
\label{dis}
We proposed a simple but effective framework for topic modelling, i.e., G2T, which incorporates contextual pretrained language models and community detection algorithms in complex networks. Experiments on both English and Chinese documents with different lengths suggested that G2T achieved state-of-the-art performance in topic modelling. Automatic evaluations reported that G2T can produce better coherent and diverse topics than baselines. Compared with clustering-based topic models using clustering methods such as k-means and HDBSCAN, G2T can make better use of contextual pretrained language models. Compared with conventional topic models such as LDA, the operating efficiency is within an acceptable scope.


\end{document}